\documentclass[conference]{IEEEtran}
\IEEEoverridecommandlockouts
\usepackage{cite}
\usepackage{amsmath,amssymb,amsfonts}
\usepackage{algorithmic}
\usepackage{graphicx}
\usepackage{textcomp}
\usepackage{xcolor}
\def\BibTeX{{\rm B\kern-.05em{\sc i\kern-.025em b}\kern-.08em
 T\kern-.1667em\lower.7ex\hbox{E}\kern-.125emX}}
\begin{document}

\title{A Multi-Model Metric-based Selection Framework for Abstractive Text summarization\\}

\author{\IEEEauthorblockN{ Ahmed Alansary}
\IEEEauthorblockA{\textit{Faculty of Computer Science} \\
\textit{MSA University}\\
Giza, Egypt \\
ahmed.mohamed406@msa.edu.eg}
\and
\IEEEauthorblockN{ Ali Hamdi}
\IEEEauthorblockA{\textit{Faculty of Computer Science} \\
\textit{MSA University}\\
Giza, Egypt \\
ahamdi@msa.edu.eg}}

\IEEEpubid{\makebox[\columnwidth]{ 979-8-3315-8488-7/26/\$31.00 ©2026 IEEE \hfill}
\hspace{\columnsep}\makebox[\columnwidth]{ }}
\maketitle
\IEEEpubidadjcol

\begin{abstract}
Automatic text summarization has become increasingly important due to the rapid growth of digital textual information. This paper presents a \textit{Multi-Model Summarization Framework} designed to improve the robustness and quality of abstractive text summarization. Relying on a single model often leads to inconsistent summarization quality across articles with varying structures and topics. To address this limitation, the proposed framework integrates multiple fine-tuned transformer-based summarization models and introduces a metric-based selection mechanism. In this framework, each model independently generates a candidate summary for the same input article. The generated summaries are then evaluated using automatic evaluation metrics that capture both lexical similarity and semantic relevance. Based on these scores, the framework selects the highest-quality summary as the final output. The models are fine-tuned and evaluated on the widely used CNN/DailyMail news summarization dataset. Experimental results demonstrate that the proposed framework achieves the highest BERTScore among all compared methods with a score of 88.63\%. It also outperforms several LLMs such as GPT3-D2, Falcon-7b, and Mpt-7b, highlighting its effectiveness and robustness. These findings highlight the effectiveness of leveraging multiple transformer-based models within a metric-based selection strategy to improve the quality and robustness of automatic text summarization systems.

\end{abstract}


\section{Introduction}
As a consequence of rapid growth of digital data, there is a tremendous amount of texts available through different media sources, social networks, and Internet databases. The ability to quickly analyze and comprehend the text is becoming a more and more pressing issue. Text summarization has become a key task for NLP, which concentrates on creating compact summaries that contain all the important information contained in the original text. News summarization is one of the most important tasks since it helps readers to get key information from long articles\cite{paper1,paper2}.

Traditional methods used in this field have mostly relied on extractive systems which focus on selecting the most useful sentences from the text. Such methods usually follow a sequence of steps including text processing, feature extraction, scoring of sentences, applying a base model, sentence selection, and generating the summary. An analysis of various methods of extractive summarization will show the variety of methods used such as statistical methods, rule-based methods, fuzzy logic, optimization methods, graph-based, clustering-based, machine learning and deep learning methods\cite{paper1}. Although certain extraction techniques have helped greatly in developing summarizers, many studies have shown that today’s systems still encounter difficulties which include lack of robustness on different articles’ structure and subject matter, poor performance when employing only one model architecture, and dependency on a single model output that does not always address the lexical and semantic aspects of good summaries\cite{paper1,paper4,paper5}.

Transformers have demonstrated impressive performance on benchmark tasks like CNN/DailyMail, due to their self-attention mechanism and ability to process sequences in parallel, which help in capturing long-distance textual relationships\cite{paper3,paper22}. In the same vein, researchers have proposed hybrid approaches for OCR and summarization using various deep learning models such as LSTM, Bidirectional LSTM, BERT, and T5\cite{paper2}. Despite all these developments, deep learning algorithms still continue to display certain limitations in terms of high computing requirements, dependency on training data quality, and their dependency on evaluation measures like ROUGE which might not be able to comprehensively measure the semantic quality of summaries\cite{paper3}.

The advent of Large Language Models (LLMs) has increased the opportunities for text summarization. In recent times, researchers have explored the use of several LLMs, using architectures like MPT-7B, Falcon-7B, and ChatGPT, to create abstractive summaries\cite{paper5}. While these examples do highlight the flexibility of modern summarizers, they do show one clear limitation that exists in many of the current models: the fact that the majority of these models rely only on one model at inference time.

Considering these constraints, this work introduces \textit{a Multi-Model Summarization Framework} for the CNN/DailyMail dataset. Instead of using a single model to generate summaries like in the previous studies, the proposed framework adopts the use of several transformer-based models for generating candidate summaries for every news article. More specifically, three different models will be used for generating alternative summaries for an article, and all the alternative summaries will be ranked according to the results generated from automatic measures such as lexical overlap, n-grams similarity, and semantics match. A combination of the above measures is used to rank alternative summaries and choose the best one for an article.

\section{Related Work}
Automatic Text Summarization (ATS) has been well-researched as one of the principal NLP problems, concerned with generating compact representations of lengthy documents while retaining the key pieces of information contained in them\cite{paper26}. With the increase in the amount of digitized text, both in news and academic articles and in social media content, there has been an increasing need for efficient text summarization tools\cite{paper7,paper8}. Early approaches to this problem have mostly focused on extractive approaches to summarization, which involve selecting relevant sentences in order to form a summary. There have been surveys of various summarization methodologies throughout the history of research in this domain, describing the basic pipeline of this process starting with document preprocessing, feature extraction and sentence scoring, all the way to the generation of summaries\cite{paper1,paper6,paper24}. Moreover, many of these surveys emphasize the growing complexity introduced to this task due to multi-document, multilingual, and multimodal settings\cite{paper1}. In addition, some surveys have traced the progress in abstractive summarization, covering the developments in model architectures, datasets, evaluation and benchmarks, such as CNN/DailyMail, and common metrics, like ROUGE\cite{paper19,paper23}. Together, these works provide a good insight into the present state of summarization research and define several open problems.

The classic approach to extractive summarization has been relying on statistics, graph structures, and handcrafted features to find relevant sentences. Graph-based models have gained quite an attention, where the sentence relevance was based on relationships between similar sentences. Thus, sentence centrality and semantic similarity were used to build graphs of relationships between textual entities and make sentence selection more efficient\cite{paper13}. Similarly, some ranking models have been used to aggregate various sentence-level features, such as topical content, semantic representations, keywords, and positional features, to determine sentence importance in the document\cite{paper12}. Some approaches have aimed to enhance traditional algorithms like TextRank with word embeddings and weightings to provide better representation of sentences and improve the output\cite{paper15}. Unsupervised approaches have investigated the application of clustering and topic modeling to reduce topical bias and achieve balanced coverage of document sub-topics\cite{paper20}. Despite being efficient and easily interpretable, extractive summarization approaches are generally prone to generating redundant output and have natural limitations with regards to generating coherent and paraphrased summary.

In order to address these issues, attention of researchers has gradually shifted towards abstractive summarization using neural architectures. Sequence-to-sequence architectures based on Recurrent Neural Networks (RNN) and (LSTM) have become widely popular for generating paraphrased summary by rephrasing the content of the source documents while keeping the meaning intact\cite{paper16}. They are normally built upon the encoder-decoder framework with attention mechanisms in order to model context dependencies between words and sentences. Their performance has been additionally improved through bidirectional encoders, deeper stacked architectures and advanced attention mechanisms to enhance the sequence modeling\cite{paper16,paper25}. Discourse-aware neural architectures have been developed to learn long-range dependencies and structural relationships between discourse units to enhance extractive summarization through incorporation of document-level structure\cite{paper10}.

More recently, Transformer architectures and pre-trained language models have demonstrated notable improvements in the summarization performance. Transformer models use self-attention in order to be able to better model long-range contextual relationships compared to previous neural approaches\cite{paper3}. Pre-trained models like PEGASUS-xsum, BART, and T5 demonstrate outstanding performance on standard benchmarks due to large-scale pre-training followed by fine-tuning on a specific task\cite{paper4,paper18}. However, fine-tuning of large pre-trained models usually poses concerns about overfitting and computational requirements, which is why the research on optimization and adaptation approaches has been done\cite{paper4}. Also, alternative formulations of the problem, such as modeling of summarization as a semantic matching problem between source documents and candidate summaries, proved to achieve competitive results on CNN/DailyMail benchmark\cite{paper17}. Frameworks for multi-modal summarization involving OCR and deep learning have expanded the scope of summarization to texts extracted from images\cite{paper2}.

Aside from general-domain applications, LLMs have been successfully utilized in specialized domains, such as clinical text summarization, where adapted models have achieved or exceeded the performance of human experts on certain tasks\cite{paper11}. Despite the great progress in deep learning and LLM-based summarization systems, the problems of factual consistency, semantic accuracy and evaluation of generated outputs remain\cite{paper18}. These limitations motivate further research in hybrid frameworks, capable of combining the advantages of different summarization models.

\begin{table*}[t]
\centering
\caption{Example sample from the CNN/DailyMail dataset.}
\label{tab:dataset_samples}
\scriptsize
\begin{tabular}{p{0.75\linewidth} p{0.15\linewidth}}
\hline
\textbf{Article} & \textbf{Highlights} \\
\hline
Liverpool target Neto is also wanted by PSG and clubs in Spain as Brendan Rodgers faces stiff competition to land the Fiorentina goalkeeper, according to the Brazilian's agent Stefano Castagna. The Reds were linked with a move for the 25-year-old, whose contract expires in June, earlier in the season when Simon Mignolet was dropped from the side. A January move for Neto never materialised but the former Atletico Paranaense keeper looks certain to leave the Florence-based club in the summer. It had been reported that Neto had a verbal agreement to join Serie A champions Juventus at the end of the season but his agent has revealed no decision about his future has been made yet. And Castagna claims Neto will have his pick of top European clubs when the transfer window re-opens in the summer, including Brendan Rodgers' side. 'There are many European clubs interested in Neto, such as for example Liverpool and Paris Saint-Germain,' Stefano Castagna is quoted as saying by Gazzetta TV. Firoentina goalkeeper Neto saves at the feet of Tottenham midfielder Nacer Chadli in the Europa League. 'In Spain too there are clubs at the very top level who are tracking him. Real Madrid? We'll see. 'We have not made a definitive decision, but in any case he will not accept another loan move elsewhere.' Neto, who represented Brazil at the London 2012 Olympics but has not featured for the senior side, was warned against joining a club as a No 2 by national coach Dunga. Neto joined Fiorentina from Atletico Paranaense in 2011 and established himself as No1 in the last two seasons. & Fiorentina goalkeeper Neto has been linked with Liverpool and Arsenal.
Neto joined Firoentina from Brazilian outfit Atletico Paranaense in 2011.
He is also wanted by PSG and Spanish clubs, according to his agent.
CLICK HERE for the latest Liverpool news. \\
\hline
\end{tabular}
\end{table*}

\section{Dataset}
The CNN/DailyMail news summarization dataset will be employed as a benchmark in this study. It is a widely-used large-scale dataset which is often used in supervised summarization studies. The dataset consists of news articles from CNN and Daily Mail website and human-made highlight sentences which serve as summaries of the news articles. In the framework of summarization, these highlight sentences are concatenated to obtain the target summary for a particular article.

The dataset consists of more than 300,000 articlesummary pairs written by professional journalists. Articles consist of 500800 words on average while the summaries are comprised of 3-5 sentences representing the main ideas of the news article. Two main attributes are present in each dataset entry \texttt{Article} and \texttt{Highlights}. The former represents the text of the news article while the latter contains a summary of the article written by an author. Some samples from the dataset are presented in Table~\ref{tab:dataset_samples}.

In accordance with the common practice, the dataset is split into three parts - training, validation and testing. The number of samples in the training part equals 287,113, in the validation part equals 13,368, and in the testing part equals 11,490.

Initially developed for machine reading comprehension and question answering purposes, CNN/DailyMail dataset has been used for abstractive summarization by using article highlights as summaries. The dataset has a large size and good quality of journalistic work which has made it one of the most popular benchmarks in summarization evaluation field\cite{paper21}.

% \begin{figure*}
%  \centering
%  \includegraphics[width=\textwidth]{fine_chart.png}
%  \caption{Visual Comparison between fine-tuned models and MASF}
%  \label{fig:base_fig}
% \end{figure*}

\section{Methodology}

This study proposes a multi-model response summarization framework designed to improve the quality of abstractive summaries by leveraging multiple transformer-based language models and automatically selecting the most informative output. Unlike traditional approaches that rely on a single summarization model, the proposed framework employs multiple fine-tuned models to generate candidate summaries and then selects the most suitable summary using an automatic evaluation mechanism.

The proposed architecture consists of five main stages: dataset preparation, model fine-tuning, input preprocessing, multi-model summarization, and Metric-based selection and evaluation.

\subsection{Model Fine-Tuning}
To improve domain adaptation and summarization performance, the pretrained models are fine-tuned on the training dataset. Three transformer-based models are used in this framework:

\begin{equation}
\mathcal{M} = \{M_{1}, M_{2}, M_{3}\}
\end{equation}

where $M_{1}$ represents T5-small, $M_{2}$ represents PEGASUS-xsum, and $M_{3}$ represents LED-base.

The T5-small and PEGASUS-xsum models are fine-tuned using a standard sequence-to-sequence training objective. During training, the article text is used as input and the reference summary is used as the target output. Tokenization is applied using the corresponding tokenizer for each model, and the models learn to generate summaries that align with the reference highlights.

For the LED-base model, parameter-efficient fine-tuning is applied using Low-Rank Adaptation (LoRA). Instead of updating all model parameters, LoRA introduces trainable low-rank matrices into specific attention layers of the transformer architecture. Let the adapted model be denoted as

\begin{equation}
M_{3}' = \text{LoRA}(M_{3})
\end{equation}

where only the injected LoRA parameters are trained while the base model parameters remain frozen. This approach reduces the number of trainable parameters while maintaining effective summarization performance.

\begin{figure}[t]
 \centering
\includegraphics[width=0.8\linewidth]{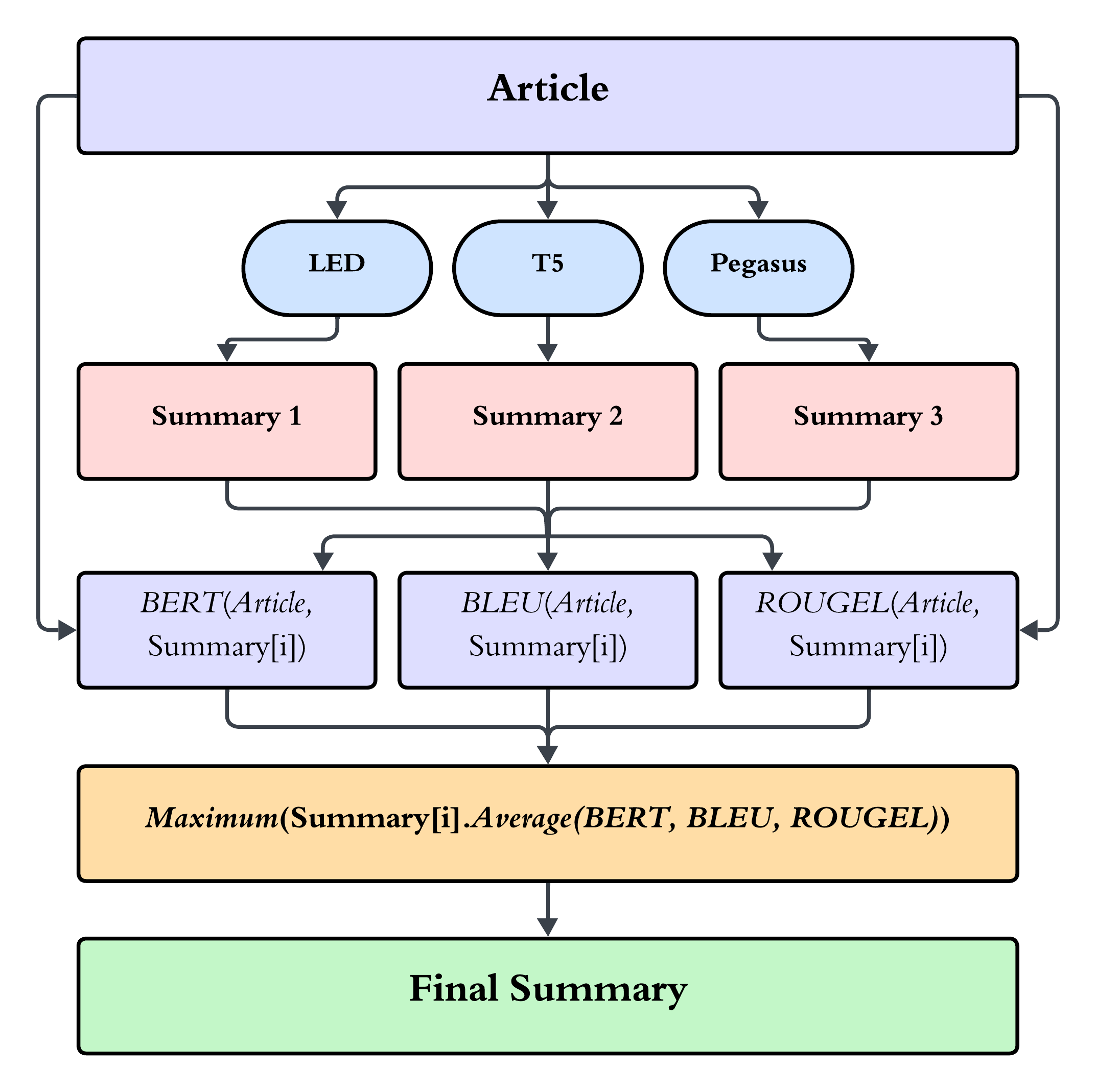}
 \caption{Overview of the proposed multi-model metric-based summarization framework.}
 \label{fig:model}
\end{figure}

\subsection{Input Preprocessing}

Given an input article $A$, the text is first tokenized and truncated to ensure compatibility with the maximum input length supported by the models. In the case of T5-small and LED-base, task-specific prompts are appended to guide the summarization process.

Let the preprocessing function be denoted as

\begin{equation}
A' = f(A)
\end{equation}

where $A'$ represents the processed input sequence that is fed into the summarization models.

\subsection{Multi-Model Summarization}

After fine-tuning, the processed article $A'$ is passed to the three summarization models to generate candidate summaries as illustrated in Figure \ref{fig:model}. Each model independently produces a summary for the same input article.

Formally, the summary generated by model $M_i$ is defined as

\begin{equation}
S_i = M_i(A')
\end{equation}

where $S_i$ denotes the candidate summary produced by model $M_i$. This process results in a set of candidate summaries:

\begin{equation}
\mathcal{S} = \{S_1, S_2, S_3\}
\end{equation}

Each summary is generated using a maximum generation length of 128 tokens.

\subsection{Automatic Evaluation}

To assess the quality of the generated summaries, three complementary evaluation metrics are used: ROUGE-L, BLEU and BERTScore. ROUGE-L measures lexical overlap between the generated summary and the article, BLEU measures the precision of n-gram overlap between the generated summary and the article, while BERTScore measures semantic similarity using contextual embeddings.

Let $R(S_i)$ denote the ROUGE-L score of summary $S_i$, $BL(S_i)$ denote the BLEU score, and $BS(S_i)$ denote the BERTScore F1 value. A combined evaluation score is computed as

\begin{equation}
 Score(S_i) = \frac{R(S_i) + BL(S_i) + BS(S_i)}{3}
\end{equation}

This combined score captures both surface-level textual overlap and deeper semantic similarity between the generated summaries and articles.

\begin{table}[b]
\centering
\renewcommand{\arraystretch}{1.3} % increase row height (default is 1.0)

\caption{Performance comparison between base-line models and \textbf{(Our)}}
\label{tab:results_base}
\begin{tabular}{lcccc}
\hline
\textbf{Model} & \textbf{BERTScore} & \textbf{ROUGE-L} & \textbf{BLEU} & \textbf{AverageScore} \\
\hline
LED-base & 86.33\% & 21.62\% & 6.36\% & 38.10\% \\
T5-small & 86.31\% & 25.19\% & 7.75\% & 39.75\% \\
Pegasus-xsum & 86.36\% & 17.03\% & 2.34\% & 35.24\% \\

\textbf{(Our)} & \textbf{87.07\%} & \textbf{24.30\%} & \textbf{7.40\%} & \textbf{39.59\%}\\

\hline
\end{tabular}
\end{table}

\subsection{Metric-based Selection}

The final stage of the framework selects the most informative summary among the candidate outputs. Let $S^{*}$ denote the final selected summary. The metric-based selection mechanism chooses the summary with the highest evaluation score:

\begin{equation}
 S^{*} = \arg\max_{S_i \in \mathcal{S}} Score(S_i)
\end{equation}

By selecting the best-performing summary from multiple fine-tuned models, the framework leverages the complementary strengths of different transformer architectures. This metric-based strategy improves robustness and increases the likelihood of producing high-quality summaries across diverse input articles.

\begin{table*}[h]
\centering
\renewcommand{\arraystretch}{1} % increase row height (default is 1.0)
\caption{Performance comparison between recent related works on CNN/DailyMail dataset and \textbf{(Our)}}
\label{tab:results_all}
\begin{tabular}{clcccc}
\hline
\textbf{Reference} & \textbf{Model} & \textbf{BERTScore} & \textbf{ROUGE-L} & \textbf{BLEU} & \textbf{AverageScore} \\
\hline

\cite{paper18} & PEGASUS-large & - & 21.73\% & 5.01\% & -\\

\cite{paper18} & T5-base & - & 23.54\% & 5.65\% & -\\

\cite{paper18} & T5-large & - & 24.42\% & 5.75\% & -\\

\cite{paper18} & BART-CNN-large & - & 24.58\% & 6.68\% & -\\

\cite{paper5} & falcon-7b-instruct & 83.8\% & 19.70\% & $9.47 \times 10^{-230}\%$ & 34.50\% \\

\cite{paper5} & mpt-7b-instruct & 85.10\% & 21.30\% & $9.35 \times 10^{-230}\%$ & 35.46\% \\

\cite{paper14} & GPT3-D2 & 85.90\% & 28.08\% & 6.60\% & 40.19\%\\

\cite{paper14} & T0 & 86.30\% & 33.95\% & 8.90\% & 43.05\%\\

\cite{paper14} & PEGASUS-xsum & 86.50\% & 36.65\% & 12.20\% & 45.11\%\\

\cite{paper14} & BRIO & 87.10\% & 39.21\% & 11.70\% & 46.00\%\\

\cite{paper5} & text-davinci-003 & 86.80\% & 25.50\% & \textbf{48.96\%} & \textbf{53.75\%} \\

\cite{paperX} & DAMB & \textbf{88.70\%} & 30.50\% & - & -\\

 & \textbf{(Our)} & 88.63\% & \textbf{32.75\%} & 16.00\% & 45.80\%\\
\hline
\end{tabular}
\end{table*}

\section{Results and Discussion}

\begin{figure}[h]
 \centering
 \includegraphics[width=\linewidth]{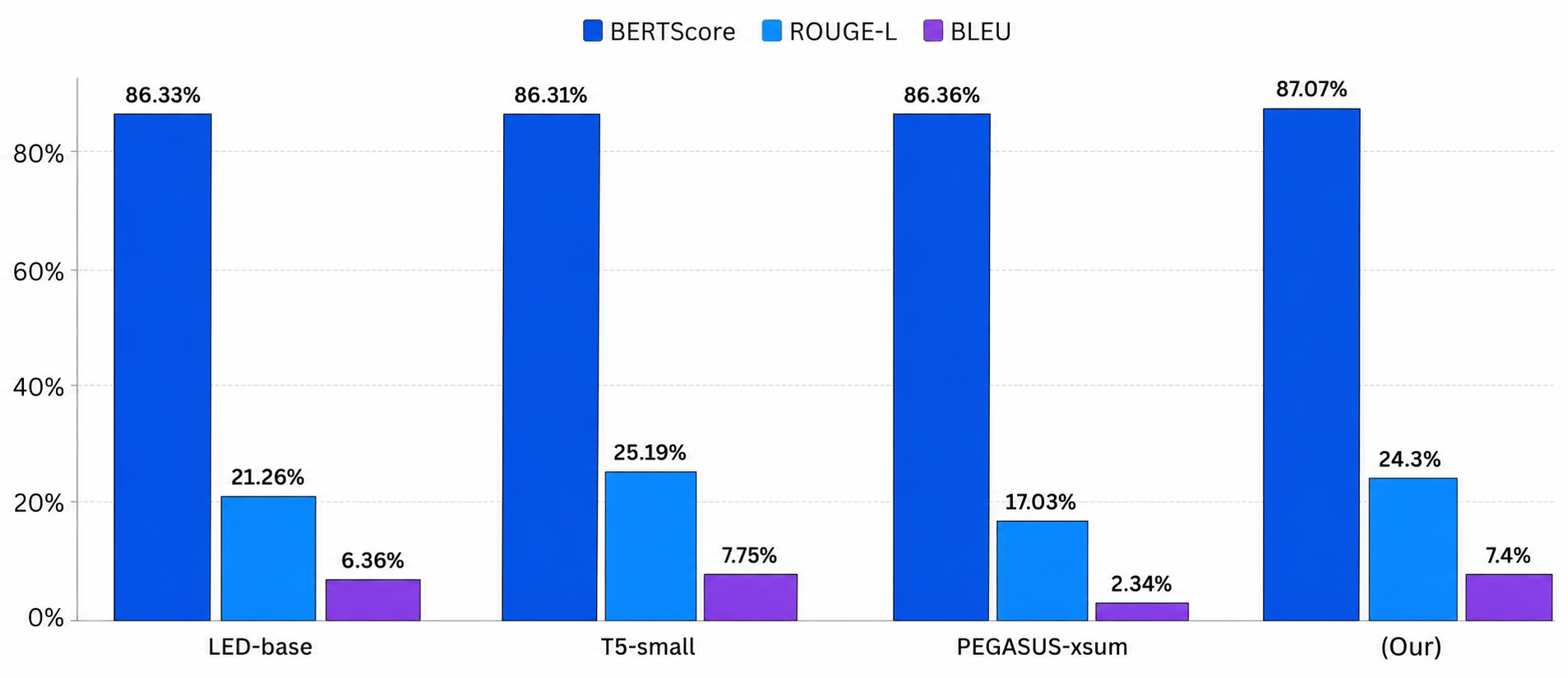}
 \caption{Visual Comparison between base-line models and (Our)}
 \label{fig:base_fig}
\end{figure}

\begin{figure}[t]
 \centering
 \includegraphics[width=\linewidth]{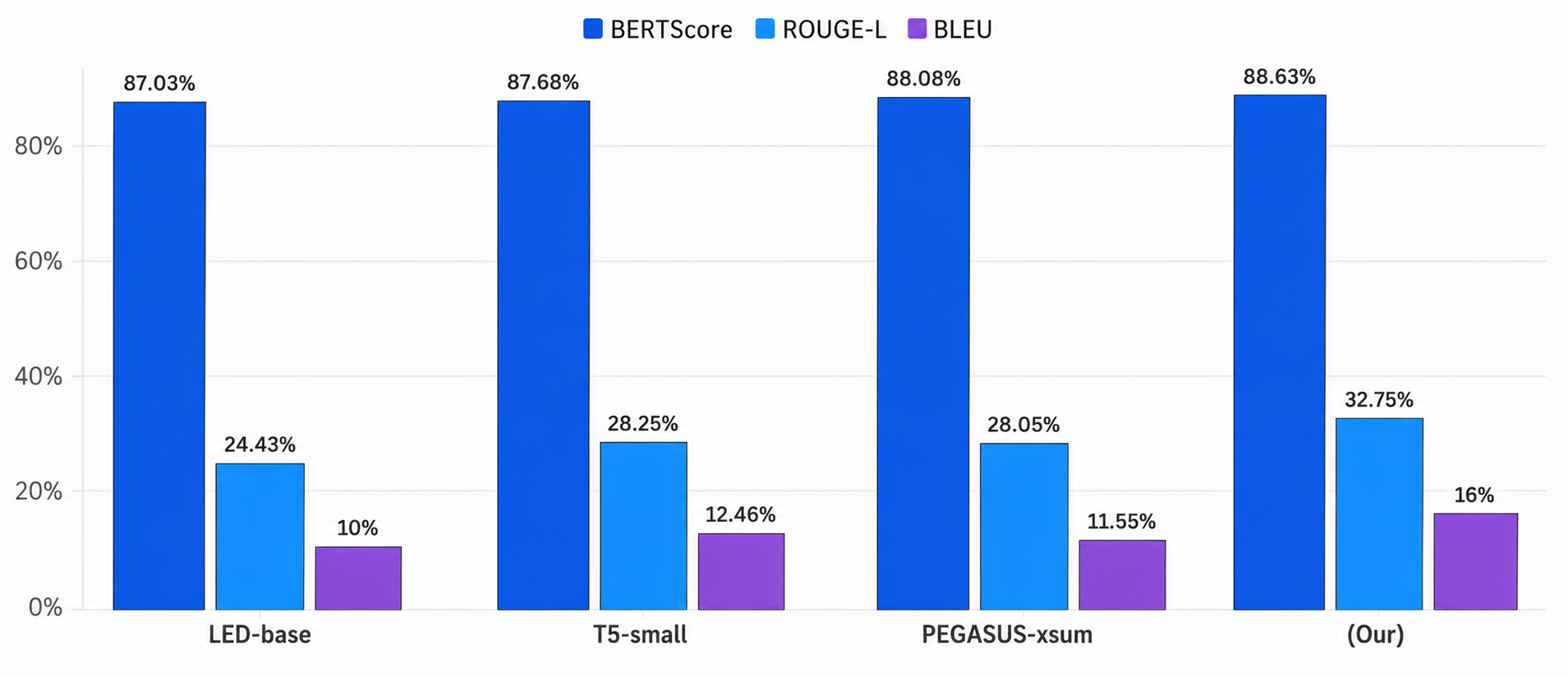}
 \caption{Visual Comparison between fine-tuned models and (Our)}
 \label{fig:fine_fig}
\end{figure}

The experimental results presented in Tables~\ref{tab:results_base},~\ref{tab:results} together with the corresponding bar charts in Figures~\ref{fig:base_fig} and~\ref{fig:fine_fig}, demonstrate the effectiveness of the proposed \textbf{Multi-Model Metric-based Summarization Framework} in comparison with individual transformer-based models.

In the base-line setting, This framework achieves the highest BERTScore of 87.07\%, outperforming LED-base (86.33\%), T5-small (86.31\%), and PEGASUS-xsum (86.36\%). Although T5-small records slightly higher ROUGE-L and BLEU scores of 25.19\% and 7.75\%, respectively, compared with 24.30\% and 7.40\% for this framework, it maintains a highly competitive overall average score of \textbf{39.59\%}, which is very close to the best-performing individual base-line model at 39.75\%. As illustrated in the base-line bar chart, the proposed framework exhibits a more balanced distribution across the evaluation metrics, reflecting stronger consistency in summary quality.

\begin{table}[b]
\centering
\renewcommand{\arraystretch}{1.3} % increase row height (default is 1.0)
\caption{Performance comparison between fine-tuned models and \textbf{(Our)}}
\label{tab:results}
\begin{tabular}{lcccc}
\hline
\textbf{Model} & \textbf{BERTScore} & \textbf{ROUGE-L} & \textbf{BLEU} & \textbf{AverageScore} \\
\hline
LED-base & 87.03\% & 24.43\% & 10.00\% & 40.48\% \\
T5-small & 87.68\% & 28.25\% & 12.46\% & 42.79\% \\
Pegasus-xsum & 88.08\% & 28.05\% & 11.55\% & 42.56\% \\
\textbf{(Our)} & \textbf{88.63\%} & \textbf{32.75\%} & \textbf{16.00\%} & \textbf{45.80\%}\\
\hline
\end{tabular}
\end{table}

A more substantial improvement is observed in the fine-tuned setting, where this framework achieves the best performance across all evaluation metrics. Specifically, it records a BERTScore of 88.63\%, a ROUGE-L score of 32.75, and a BLEU score of 16.00\%, resulting in the highest average score of 45.80\%. Compared with the strongest individual fine-tuned model, T5-small, which achieves an average score of 42.79\%, the proposed framework improves the overall performance by 3.01\%. This improvement is particularly evident in the BLEU metric, where this framework significantly outperforms PEGASUS-xsum (11.55\%), T5-small (12.46\%), and LED-base (10.00\%). The corresponding bar chart further reinforces this result by showing that this framework spans the largest area among all compared models, indicating superior balance between semantic similarity and lexical alignment.

When compared with recent related works, as shown in Table~\ref{tab:results_all}, the proposed framework achieves a BERTScore of 88.63\%, which is higher than all reported models in the comparison, including BRIO (87.10\%) and text-davinci-003 (86.80\%). In terms of the overall average score, this framework achieves 45.80\%, outperforming several strong recent approaches such as GPT3-D2 (40.19\%) and T0 (43.05\%), while remaining close to PEGASUS-xsum from~\cite{paper14} (45.11\%) and BRIO (46.00\%). Although text-davinci-003 reports a higher average score of 53.75\%, the proposed framework demonstrates stronger consistency across all three evaluation metrics without relying on a large model.

Overall, the results confirm that the proposed framework provides consistently strong summarization performance, with the most notable gains observed after fine-tuning. The framework not only surpasses the individual constituent models but also remains highly competitive with recent approaches, demonstrating its effectiveness in generating robust and high-quality abstractive summaries against Large Language Models (LLMs).

\section{Conclusion}
The presented approach is based on a multi-model summarization framework with the goal of improving the efficiency of news text summarization of the CNN/DailyMail dataset. Contrary to traditional solutions using only one summarization model, this framework involves multiple transformer-based models that produce multiple candidate summaries for an input text. In turn, the automatic evaluation tool uses ROUGE-L, BLEU, and BERTScore for determining the most informative summary using metric-based selection approach.

Experimental results prove the superiority of the developed framework to any baseline or fine-tuned summarization model in terms of summarization quality. As a result of leveraging complementary advantages of different transformer architectures, the framework is able to provide more consistent and reliable summaries for diverse input texts. The use of metric-based selection tool makes it possible to automatically choose the most relevant summary for every input text that improves the robustness of the framework. Also, compared to some recent works, the proposed framework outperforms several large models, like GPT3-D2 and T0. Besides, the framework performs much better than other competitive models, such as PEGASUS-xsum and BRIO. It reaches the highest BERTScore in comparison with all other methods considered which means that summaries produced by this framework have high semantic consistency.

For future work, it can be useful to consider additional evaluation metrics, namely, reference-free and human-aligned evaluation methods for the improvement of the selection process. Also, it can be interesting to explore the possibility of adding more large language models as well as model adaptation techniques.

\bibliographystyle{IEEEtran}
\bibliography{ref}

@inproceedings{paper1,
  title={Summn: A multi-stage summarization framework for long input dialogues and documents},
  author={Zhang, Yusen and Ni, Ansong and Mao, Ziming and Wu, Chen Henry and Zhu, Chenguang and Deb, Budhaditya and Awadallah, Ahmed and Radev, Dragomir and Zhang, Rui},
  booktitle={Proceedings of the 60th Annual Meeting of the Association for Computational Linguistics (Volume 1: Long Papers)},
  pages={1592--1604},
  year={2022}
}

@inproceedings{paper2,
  title={ChatGPT vs human-authored text: Insights into controllable text summarization and sentence style transfer},
  author={Liu, Dongqi and Demberg, Vera},
  booktitle={Proceedings of the 61st Annual Meeting of the Association for Computational Linguistics (Volume 4: Student Research Workshop)},
  pages={1--18},
  year={2023}
}

@article{paper3,
    author = {Rennard, Virgile and Shang, Guokan and Hunter, Julie and Vazirgiannis, Michalis},
    title = {Abstractive Meeting Summarization: A Survey},
    journal = {Transactions of the Association for Computational Linguistics},
    volume = {11},
    pages = {861-884},
    year = {2023},
    month = {07},
    issn = {2307-387X},
    doi = {10.1162/tacl_a_00578}
}

@article{paper4,
  title={Domain adaptation with pre-trained transformers for query-focused abstractive text summarization},
  author={Laskar, Md Tahmid Rahman and Hoque, Enamul and Huang, Jimmy Xiangji},
  journal={Computational Linguistics},
  volume={48},
  number={2},
  pages={279--320},
  year={2022},
  publisher={MIT Press One Broadway, 12th Floor, Cambridge, Massachusetts 02142, USA~…}
}

@misc{paper5,
      title={Text Summarization Using Large Language Models: A Comparative Study of MPT-7b-instruct, Falcon-7b-instruct, and OpenAI Chat-GPT Models}, 
      author={Lochan Basyal and Mihir Sanghvi},
      year={2023},
      eprint={2310.10449},
      archivePrefix={arXiv},
      primaryClass={cs.CL},
      url={https://arxiv.org/abs/2310.10449}, 
}

@inproceedings{paper6,
    title = "A Survey of Automatic Text Summarization Using Graph Neural Networks",
    author = "Salchner, Marco Ferdinand  and
      Jatowt, Adam",
    editor = "Calzolari, Nicoletta  and
      Huang, Chu-Ren  and
      Kim, Hansaem  and
      Pustejovsky, James  and
      Wanner, Leo  and
      Choi, Key-Sun  and
      Ryu, Pum-Mo  and
      Chen, Hsin-Hsi  and
      Donatelli, Lucia  and
      Ji, Heng  and
      Kurohashi, Sadao  and
      Paggio, Patrizia  and
      Xue, Nianwen  and
      Kim, Seokhwan  and
      Hahm, Younggyun  and
      He, Zhong  and
      Lee, Tony Kyungil  and
      Santus, Enrico  and
      Bond, Francis  and
      Na, Seung-Hoon",
    booktitle = "Proceedings of the 29th International Conference on Computational Linguistics",
    month = oct,
    year = "2022",
    address = "Gyeongju, Republic of Korea",
    publisher = "International Committee on Computational Linguistics",
    url = "https://aclanthology.org/2022.coling-1.536/",
    pages = "6139--6150"
}

@article{paper7,
  title={Automatic text summarization: A comprehensive survey},
  author={El-Kassas, Wafaa S and Salama, Cherif R and Rafea, Ahmed A and Mohamed, Hoda K},
  journal={Expert systems with applications},
  volume={165},
  pages={113679},
  year={2021},
  publisher={Elsevier}
}

@article{paper8,
  title={A comprehensive survey on automatic text summarization with exploration of LLM-based methods},
  author={Zhang, Yang and Jin, Hanlei and Meng, Dan and Wang, Jun and Tan, Jinghua},
  journal={Neurocomputing},
  pages={131928},
  year={2025},
  publisher={Elsevier}
}

@inproceedings{paper10,
    title = "Discourse-Aware Neural Extractive Text Summarization",
    author = "Xu, Jiacheng  and
      Gan, Zhe  and
      Cheng, Yu  and
      Liu, Jingjing",
    editor = "Jurafsky, Dan  and
      Chai, Joyce  and
      Schluter, Natalie  and
      Tetreault, Joel",
    booktitle = "Proceedings of the 58th Annual Meeting of the Association for Computational Linguistics",
    month = jul,
    year = "2020",
    address = "Online",
    publisher = "Association for Computational Linguistics",
    url = "https://aclanthology.org/2020.acl-main.451/",
    doi = "10.18653/v1/2020.acl-main.451",
    pages = "5021--5031"
}

@article{paper11,
  title={Adapted large language models can outperform medical experts in clinical text summarization},
  author={Van Veen, Dave and Van Uden, Cara and Blankemeier, Louis and Delbrouck, Jean-Benoit and Aali, Asad and Bluethgen, Christian and Pareek, Anuj and Polacin, Malgorzata and Reis, Eduardo Pontes and Seehofnerov{\'a}, Anna and others},
  journal={Nature medicine},
  volume={30},
  number={4},
  pages={1134--1142},
  year={2024},
  publisher={Nature Publishing Group US New York}
}

@article{paper12,
  title={RankSum—An unsupervised extractive text summarization based on rank fusion},
  author={Joshi, Akanksha and Fidalgo, Eduardo and Alegre, Enrique and Alaiz-Rodriguez, Rocio},
  journal={Expert Systems with Applications},
  volume={200},
  pages={116846},
  year={2022},
  publisher={Elsevier}
}

@INPROCEEDINGS{paper13,
  author={Jain, Minni and Rastogi, Harshita},
  booktitle={2020 4th International Conference on Electronics, Communication and Aerospace Technology (ICECA)}, 
  title={Automatic Text Summarization using Soft-Cosine Similarity and Centrality Measures}, 
  year={2020},
  volume={},
  number={},
  pages={1021-1028},
  doi={10.1109/ICECA49313.2020.9297583}}

@misc{paper14,
      title={News Summarization and Evaluation in the Era of GPT-3}, 
      author={Tanya Goyal and Junyi Jessy Li and Greg Durrett},
      year={2023},
      eprint={2209.12356},
      archivePrefix={arXiv},
      primaryClass={cs.CL},
      url={https://arxiv.org/abs/2209.12356}, 
}

@inproceedings{paper15,
    title = "Extractive Summarization using Extended {T}ext{R}ank Algorithm",
    author = "N. Vora, Ansh  and
      Jain, Rinit Mayur  and
      Shah, Aastha Sanjeev  and
      Sonawane, Sheetal",
    editor = "Lalitha Devi, Sobha  and
      Arora, Karunesh",
    booktitle = "Proceedings of the 21st International Conference on Natural Language Processing (ICON)",
    month = dec,
    year = "2024",
    address = "AU-KBC Research Centre, Chennai, India",
    publisher = "NLP Association of India (NLPAI)",
    pages = "462--471"
}

@inproceedings{paper16,
  title={Bidirectional LSTM networks for abstractive text summarization},
  author={Kova{\v{c}}evi{\'c}, Aldin and Ke{\v{c}}o, Dino},
  booktitle={International Symposium on Innovative and Interdisciplinary Applications of Advanced Technologies},
  pages={281--293},
  year={2021},
  organization={Springer}
}

@inproceedings{paper17,
    title = "Extractive Summarization as Text Matching",
    author = "Zhong, Ming  and
      Liu, Pengfei  and
      Chen, Yiran  and
      Wang, Danqing  and
      Qiu, Xipeng  and
      Huang, Xuanjing",
    editor = "Jurafsky, Dan  and
      Chai, Joyce  and
      Schluter, Natalie  and
      Tetreault, Joel",
    booktitle = "Proceedings of the 58th Annual Meeting of the Association for Computational Linguistics",
    month = jul,
    year = "2020",
    address = "Online",
    publisher = "Association for Computational Linguistics",
    url = "https://aclanthology.org/2020.acl-main.552/",
    doi = "10.18653/v1/2020.acl-main.552",
    pages = "6197--6208"
}

@article{paper18,
  title={Deep learning for text summarization using NLP for automated news digest},
  author={Rani Krishna, KM and Somasundaram, K and Arulmozhivarman, P and Immanuel, Sarah A and Rajkumar, ER},
  journal={Scientific Reports},
  volume={15},
  number={1},
  pages={36343},
  year={2025},
  publisher={Nature Publishing Group UK London}
}

@article{paper19,
title = {Abstractive text summarization: State of the art, challenges, and improvements},
journal = {Neurocomputing},
volume = {603},
pages = {128255},
year = {2024},
issn = {0925-2312},
doi = {https://doi.org/10.1016/j.neucom.2024.128255},
author = {Hassan Shakil and Ahmad Farooq and Jugal Kalita}
}

@article{paper20,
  title={A topic modeled unsupervised approach to single document extractive text summarization},
  author={Srivastava, Ridam and Singh, Prabhav and Rana, KPS and Kumar, Vineet},
  journal={Knowledge-Based Systems},
  volume={246},
  pages={108636},
  year={2022},
  publisher={Elsevier}
}

@article{paper21,
  title={A survey on cross-lingual summarization},
  author={Wang, Jiaan and Meng, Fandong and Zheng, Duo and Liang, Yunlong and Li, Zhixu and Qu, Jianfeng and Zhou, Jie},
  journal={Transactions of the Association for Computational Linguistics},
  volume={10},
  pages={1304--1323},
  year={2022},
  publisher={MIT Press One Broadway, 12th Floor, Cambridge, Massachusetts 02142, USA~…}
}

@inproceedings{paper22,
    title = "Anlirika: An {LSTM}{--}{CNN} Flow Twister for Spoken Language Identification",
    author = "Scherbakov, Andreas  and
      Whittle, Liam  and
      Kumar, Ritesh  and
      Singh, Siddharth  and
      Coleman, Matthew  and
      Vylomova, Ekaterina",
    editor = {Vylomova, Ekaterina  and
      Salesky, Elizabeth  and
      Mielke, Sabrina  and
      Lapesa, Gabriella  and
      Kumar, Ritesh  and
      Hammarstr{\"o}m, Harald  and
      Vuli{\'c}, Ivan  and
      Korhonen, Anna  and
      Reichart, Roi  and
      Ponti, Edoardo Maria  and
      Cotterell, Ryan},
    booktitle = "Proceedings of the Third Workshop on Computational Typology and Multilingual NLP",
    month = jun,
    year = "2021",
    address = "Online",
    publisher = "Association for Computational Linguistics",
    doi = "10.18653/v1/2021.sigtyp-1.14",
    pages = "145--148"
}

@article{paper23,
title = {LexiSem: A re-ranker balancing lexical and semantic quality for enhanced abstractive summarization},
journal = {Neurocomputing},
volume = {650},
pages = {130816},
year = {2025},
issn = {0925-2312},
doi = {https://doi.org/10.1016/j.neucom.2025.130816},
author = {Eman Aloraini and Hozaifa Kassab and Ali Hamdi and Khaled Shaban}
}

@inproceedings{paper24,
  title={Balancing Factual Consistency and Diversity in Abstractive Summarization via Model-Agnostic Composite Reranking},
  author={Elewa, Mariam and Hamdi, Ali and Kassab, Hozaifa and Shaban, Khaled},
  booktitle={2025 IEEE/ACS 22nd International Conference on Computer Systems and Applications (AICCSA)},
  pages={1--8},
  year={2025},
  organization={IEEE}
}

@inproceedings{paperX,
author = {Burukanli, Mehmet and Ari, Davut},
year = {2025},
month = {12},
pages = {},
title = {DAMB: A DYNAMIC ADAPTIVE MULTI-MODEL BENCHMARKING FRAMEWORK FOR ABSTRACTIVE TEXT SUMMARIZATION}
}

@article{paper25,
  title={SummFactScore: A Claim-Centric Framework forReference-Free Factual Consistency Evaluation inLong-Document Summarization},
  author={Aloraini, Eman and Hamdi, Ali and Elmahjub, Ezieddin and others},
  journal={Ali and Elmahjub, Ezieddin, SummFactScore: A Claim-Centric Framework forReference-Free Factual Consistency Evaluation inLong-Document Summarization}
}

@inproceedings{paper26,
  title={Riro: Reshaping inputs, refining outputs unlocking the potential of large language models in data-scarce contexts},
  author={Hamdi, Ali and Kassab, Hozaifa and Bahaa, Mohamed and Mohamed, Marwa},
  booktitle={The International Conference of Advanced Computing and Informatics},
  pages={69--79},
  year={2024},
  organization={Springer Nature Switzerland Cham}
}

\end{document}